\pdfoutput=1

\documentclass[11pt]{article}

\usepackage{acl}

\usepackage{times}
\usepackage{latexsym}

\usepackage{graphicx}
\usepackage{hyperref}

\usepackage[T1]{fontenc}

\usepackage[utf8]{inputenc}

\usepackage{microtype}
\usepackage[symbol]{footmisc}

\title{Multilinguals at SemEval-2022 Task 11: Transformer Based Architecture for Complex NER}

\author{Amit Pandey*, Swayatta Daw*, \and Vikram Pudi \\
        International Institute of Information Technology, Hyderabad, India\\
        \texttt{\{amit.pandey, swayatta.daw\}@research.iiit.ac.in}\\ \texttt{vikram@iiit.ac.in}}

\begin{document}
\maketitle
\begin{abstract}
We investigate the task of complex NER for the English language. The task is non-trivial due to the semantic ambiguity of the textual structure and the rarity of occurrence of such entities in the prevalent literature. Using pre-trained language models such as BERT, we obtain a competitive performance on this task. We qualitatively analyze the performance of multiple architectures for this task. All our models are able to outperform the baseline by a significant margin. Our best performing model beats the baseline F1-score by over 9\%. 
\footnotetext[1]{Equal Contribution}
\end{abstract}

\section{Introduction}
The Named Entity Recognition (NER) task aims to detect entities from unstructured text and classify them into predefined categories. Although the task of NER has been investigated adequately by previous research work \cite{mansouri2008named, nadeau2007survey, lample2016neural, florian2003named, ritter2011named}, the detection of named entities in open-domain settings is non-trivial. Moreover, the introduction of additional layers of complexity, in the form of semantic ambiguity and a lower amount of contextual availability, poses further challenges. 
For example, in a low-context and semantically ambiguous sentence such as \textit{Let us play Among Us}, the token sequence \textit{Among Us}, can refer to a common phrase or a popular video game, and hence be categorized as a Creative Work (CW).

Recently, deep learning models have gained popularity for NER \cite{yadav2019survey, li2020survey, habibi2017deep}. However, these approaches are data-intensive and become ineffective when there is a lack of labeled data. To foster research in this area, \cite{multiconer-report} has introduced the SemEval MultiCoNER shared task that deals with multilingual complex named entity recognition.
 
This paper describes our approach to tackle complex NER task for the English language using state-of-the-art deep learning models and introduces a simple neural network architecture that builds on top of pre-trained language models. We compare multiple architectures on the validation and test set of the shared task. All our models outperform the baseline by a significant margin. Through our experiments, we discover that leveraging transformer models based on attention mechanism~\cite{vaswani2017attention} results in better performance even in low context and ambiguous settings. The code is available at \url{https://github.com/AmitPandey-Research/Complex_NER}

We describe the prior research work done with respect to both general and low-resource NER tasks in Section \ref{Related Work}. We provide the formal task description in Section \ref{taskdescription}, the dataset details in Section \ref{dataset}, the method and the model architecture in Section \ref{System Overview}. We provide details about the experimental implementation in Section \ref{implementationdetails}. We discuss the results obtained and error analysis in Sections \ref{results} and \ref{erroranalysis} respectively, and finally, we conclude the paper in Section \ref{conclusion}.

\section{Related Work}
\label{Related Work}

A widely used benchmark for NER was the CoNLL 2003 shared task. It contained annotated newswire text from the Reuters RCV1 corpus. Previous researchers \cite{baevski-etal-2019-cloze} had used BiLSTM models with attention to predict named entities on this dataset. \cite{ma-hovy-2016-end} used a BiLSTM-CNN-CRF to predict the named entities. 

\textbf{Sequence labeling for Named Entity Recognition:} Recent approaches have aimed at utilizing deep learning techniques for training NER models. However, these techniques require a large amount of token-level labeled data for NER tasks. Annotation for such kinds of labeled datasets can be expensive, time-consuming,  and laborious. The datasets introduced in this task encompass a large number of low-resource and complex NER entities.

Recent work on NER in scientific documents has been concentrated around detecting biomedical named entities \cite{biomed} or scientific entities like tasks, methods and datasets \cite{luan-etal-2018-multi,jain-etal-2020-scirex,mesbah2018tse}. 

NER has been traditonally modelled as a sequence labelling task, using CRF \cite{Lafferty2001ConditionalRF} to classify the labels. Recent approaches have used deep learning based models \cite{Li2018ASO}. These approaches are data intensive in nature. To tackle the label scarcity problem, methods like Distant Supervision \cite{9378052,DBLP:journals/corr/abs-2006-15509,DBLP:journals/corr/abs-2102-13129}, Active Learning \cite{10.1145/3012003},  Reinforcement Learning-based Distant Supervision~\cite{nooralahzadeh-etal-2019-reinforcement,yang-etal-2018-distantly} have been proposed. 


\section{Task Description}
\label{taskdescription}
The objective of this shared task is to build complex Named Entity Recognition systems. The task presents a unique challenge in the form of detecting the entities in semantically ambiguous and low-context settings. Moreover, the shared task also tests the generalization capability and domain adaptability of the proposed systems by testing the system over additional (low-context) datasets containing questions and short search queries, such as Google Search queries.
\begin{table}[h]
    \centering
    \begin{tabular}{c|c}
          \textbf{Label} & \textbf{Description} \\
          \hline
         PER & Person \\
         LOC & Location \\
         GRP & Group \\
         CORP & Corporation \\
         PROD & Product \\
         CW & Creative Work \\
    \end{tabular}
    \caption{Entity types in the label space}
    \label{tab:stats2}
\end{table}

For this task, given an input sentence (an arbitrary sequence of tokens), the systems have to identify the B-I-O format \cite{ramshaw1999text} (short for beginning, inside, outside) tags for 6 NER entity classes: Person, Product, Location, Group, Corporation, and Creative Work. The description attributed to each class label is described in Table \ref{tab:stats2}.

\section{Dataset}
\label{dataset}
The MultiCoNER dataset~\cite{multiconer-data} consists of labeled complex Named Entities (NE).
For the monolingual track, the participants have to train a model that works for a single language. For training and validation purposes, train and dev sets are provided with labeled entities. The monolingual model trained needs to be used for the prediction of named entities in the test set that consists of more than 150K instances. The labels from the test set are not provided directly. In this system description for the monolingual track, we have considered the English NER dataset for our task. The dataset follows a BIO tagging scheme, and there are six entity types in the label space. The statistics for the English dataset in the monolingual track for the train and dev set are provided in Table \ref{tab:stats1}.
\begin{table}[h]
    \centering
    \begin{tabular}{c|c|c}
          & \textbf{Train} & \textbf{Dev} \\
          \hline
         \# sentences & 15300 & 800 \\
    \end{tabular}
    \caption{Total sentences in English monolingual track}
    \label{tab:stats1}
    \centering
\end{table}

\section{System Overview}
\label{System Overview}
This section describes our approach to designing a system to solve the problem of classifying the tokens (words) of a given sentence into one of the six NE categories.  We also briefly describe features of the BERT (Bidirectional Encoder Representations from Transformers) ~\cite{devlin2019bert} model employed in our system. 


We designed three architectures based on pre-trained language model BERT: 1) BERT+Linear, 2) BERT+CRF, and 3) BERT+BiLSTM+CRF.
A detailed explanation of these architectures is as follows:


\subsection{BERT+Linear}
We model this task as a multiclass classification problem. The first step to finding labels for the entities is to find dense vector representations of the tokens in the given sentence. 

Instead of using static pre-trained word embeddings, such as Word2Vec~\cite{mikolov2013efficient} and GloVe~\cite{pennington-etal-2014-glove} that rely only on static global representations of word vectors, we employ BERT-based context-aware representations (BERT embeddings) that leverage the full context of the entire sentence.

This helps in extracting more information for the task of NER that is highly dependent on the inter-token relationship. BERT learns the representations for the tokens in the given text by jointly considering both the left and right context of the tokens at each layer~\cite{devlin2019bert}. To better learn the inter-token dependencies, BERT leverages the attention mechanism with multiple attention heads that focus on different aspects of a token’s relation to other tokens. For an $i$th token $x_{i}$ among a sequence of tokens $x = (x_{1},x_{2},x_{3},...,x_{m})$, we obtain a low-dimensional BERT embedding, $\tilde{x_{i}} \in R^{d}$ where $d$ is the embedding dimension.

\begin{figure}[t!]
  \centering
  \includegraphics[width=0.95\linewidth]{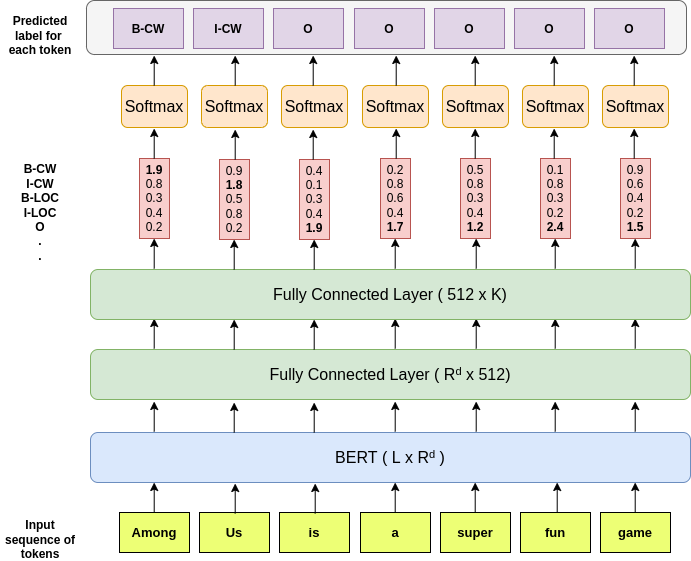}
  \caption{BERT-Linear architecture}
  \label{fig:Bert-Linear architecture}
\end{figure}

We pass this BERT token embedding to a dense classifier that consists of two fully connected layers. This classifier layer maps the BERT embeddings to lower dimension logit vectors $\tilde{x_{i}} \in R^{k}$, where $k$ is the total number of labels. The logits are then passed to the softmax normalization function. The softmax generates a probability distribution across all labels for each token, which is then used to predict the most probable label. The system architecture details are shown in Figure \ref{fig:Bert-Linear architecture}.

\subsection{BERT+CRF} We use a pre-trained BERT model to obtain the token embeddings. These embeddings are passed to a token-level classifier followed by a Linear-Chain CRF. The CRF learns the transfer rules between adjacent entity labels and returns likelihood for a sequence of labels. More formally: 1) For a sequence of tokens $x = (x_{1},x_{2},x_{3},...,x_{m})$, where $x_{i}$ is the $i$th  token among the sequence of tokens, we obtain a low-dimensional dense embedding, $\tilde{x_{i}} \in R^{d}$ where $d$ is the embedding dimension. 2) This embedding is mapped to a lower dimensional space $\tilde{x_{i}} \in R^{k}$ where $k$ is the total number of labels. 3) The output emission scores from the linear layer are obtained as $P \in R^{m\times{k}}$, where $m$ is the number of tokens. These scores are passed to the CRF layer, whose parameters are $A \in R^{k+2\times{k+2}}$. Each element $A_{ij}$ signifies the transition score from the $i$th label to the $j$th label. The 2 additional states in $A$ are the start and the end state of a sequence. For a series of tokens $x = (x_{1},x_{2},x_{3},...,x_{m})$, we obtain a series of predictions $y = (y_{1},y_{2},y_{3},...,y_{m})$. 
As described in \cite{lample-etal-2016-neural}, the score of the entire sequence is defined as :

\[s(x,y) = \sum_{i=0}^{m}A_{y_{i},y_{i+1}} + \sum_{i=1}^{m}P_{i,y_{i}}\]

The model is trained to maximize the log probability of the correct label sequence:

\[\log(p(y|x))= s(x,y)-\log(\sum_{\tilde{y}\in \textbf{$Y_X$}}e^{s(x,\tilde{y})})\]

where \textbf{$Y_X$} are all possible label sequences.

\begin{table*}[t]
    \centering
    \begin{tabular}{c|c|c|c|c|c|c|c|c|c|c|}
          & \multicolumn{3}{|c|}{\textbf{BERT+Linear}} & \multicolumn{3}{|c|}{\textbf{BERT+CRF}}& \multicolumn{3}{|c|}{\textbf{BERT+BiLSTM+CRF}}\\
         \hline
    Class Label & Prec & Rec & F1 & Prec & Rec & F1 & Prec & Rec & F1\\
    \hline
    LOC & 0.9304 & 0.9145 & \textbf{0.9224} & 0.903 & 0.9145 & 0.9087 & 0.9025 & 0.9103 & 0.9064\\
    PER & 0.9659 & 0.9759 & \textbf{0.9708} & 0.936 & 0.9586 & 0.9472 & 0.8882 & 0.9586 & 0.9221\\
    PROD & 0.7365 & 0.8367 & 0.7834 & 0.7785 & 0.7891 & \textbf{0.7838} & 0.7372 & 0.7823 & 0.7591\\
    GRP & 0.8923 & 0.9158 & \textbf{0.9039} & 0.8341 & 0.9000 &   0.8658 & 0.8466 & 0.8421 & 0.8443\\
    CW & 0.7955 & 0.7955 & \textbf{0.7955} & 0.7963 & 0.733 & 0.7633 & 0.7353 & 0.7102 & 0.7225\\
    CORP & 0.893 & 0.8653 & \textbf{0.8789} & 0.8877 & 0.8601 & 0.8737 & 0.8837 & 0.7876 & 0.8329\\
    \hline
    Average & 0.8689 & 0.8839 & \textbf{0.8758} & 0.8559 & 0.8592 & 0.8571 & 0.8322 & 0.8318 & 0.8312\\
    \end{tabular}
    \caption{Results of our models on validation dataset}
    \label{tab:results1}
\end{table*}

\begin{table}[h]
    \begin{tabular}{p{3.5 cm}|p{0.75cm}|p{0.75cm}|p{0.75cm}}
         & \textbf{Prec} & \textbf{Rec} & \textbf{F1}\\
        \hline
        Baseline System & 0.773 & 0.780 & 0.776\\
        BERT + CRF & 0.855 & 0.859 & 0.857\\
        BERT+BiLSTM+CRF & 0.832 & 0.831 & 0.831\\
        BERT + Linear & \textbf{0.868} & \textbf{0.883} & \textbf{0.875}\\
    \end{tabular}
    \caption{Comparison of model performances with baseline on validation dataset} 
    \label{tab:results2}
\end{table}

\begin{table}[h]
    \centering
    \begin{tabular}{c|c|c|c}
          & \multicolumn{3}{|c}{\textbf{BERT+Linear}}\\
         \hline
    Class Label & Prec & Rec & F1 \\
    \hline
    LOC & 0.7292 & 0.7614 & 0.7449\\
    PER & 0.8776 & 0.8922 & 0.8848\\
    PROD & 0.7079 & 0.6460 & 0.6755\\
    GRP & 0.7699 & 0.6600 & 0.7107\\
    CW & 0.5527 & 0.6299 & 0.5888\\
    CORP & 0.7253 & 0.6759 & 0.6998\\ 
    \hline
    Average & 0.7271 & 0.7109 & 0.7174\\
    \end{tabular}
    \caption{Performance of model on test dataset}
    \label{tab:results3}
\end{table}
\subsection{BERT+BiLSTM+CRF} We use a pre-trained BERT model to obtain the contextual token embeddings for the input sentence. These BERT embeddings are passed to the BiLSTM layer, where the BiLSTM layer captures the information into a hidden state representation. This representation is passed to a CRF layer that obtains the probability distributions across the sequences of labels.
Specifically, the fine-tuned BERT language model is used to map the tokens in each sentence to a distributed representation. This is used as the word embedding layer for the BiLSTM+CRF model. The BiLSTM+CRF layer is used to sequence label the sentence, and the predicted labels are obtained. The supervised learning algorithm iterates to improve its predicted label accuracy over every iteration. More formally, the process can be described as follows :
1) The target sentence comprising of $m$ tokens, is represented as $x = (x_{1},x_{2},x_{3},...,x_{m})$, where $x_{i}$ represents the $i$th token of the entire target sentence. 2) $x_{i}$ is mapped to a low dimensional dense vector, $\tilde{x_{i}} \in R^{d}$ using the pretrained BERT embeddings, where $d$ is the dimension of dense embedding. 3) The sequence of vectors is taken as an input to the BiLSTM in each time step, and the forward hidden states $\overrightarrow{h_{f}} = (\overrightarrow{h_{1}},\overrightarrow{h_{2}},\overrightarrow{h_{3}},...,\overrightarrow{h_{m}})$ and the backward hidden states $\overleftarrow{h_{b}} = (\overleftarrow{h_{1}},\overleftarrow{h_{2}},\overleftarrow{h_{3}},...,\overleftarrow{h_{m}})$ are concatenated to form the combined hidden state representation $h = [\overrightarrow{h_{f}},\overleftarrow{h_{b}}]$. 4) The combined hidden state representation $h \in R^{m\times{n}}$, where $n$ is the total size of BiLSTM, is reduced to a $k$ dimensions using a linear layer, where $k$ is the number of labels to distribute the probabilities across. 4) Finally, the CRF layer is used to obtain the probability of label sequence.

\section{Implementation Details}
\label{implementationdetails}
We implement all our transformer-based models using Pytorch and Huggingface library. We implement 3 models: 1) BERT+Linear, 2) BERT+CRF, and 3) BERT+BiLSTM+CRF. We also experiment with feature engineering by concatenating label encoded Part-of-Speech (POS) tags to the token embeddings. 
We use a dropout from 0.2 to 0.5 in all models and find that a dropout probability of 0.3 gives the best results throughout. 

In the BERT+Linear model, we use two fully connected dense linear layers as a classifier on top of the BERT embedding layer. We add a softmax layer to obtain the probability distribution across all the labels. For the BERT+Linear model, we run our experiments across 1-20 epochs. We find that the model starts to overfit after 10 epochs, and the best results are obtained after 5 epochs of training. We further experiment with BERT-base (12 attention heads) and BERT-large (16 attention heads).

For BERT+CRF and BERT+BiLSTM+CRF, we experiment across 1-100 epochs. We find that the models give the most optimal result at the 20th epoch, after which they start to overfit. We use a learning rate of ${1e^{-6}}$ for all the models. We validate the results of all models using our dev set and then use the best performing model for final evaluation on the blind test set.

\section{Results}
\label{results}

We compare the performance of our models in the validation set against the baseline. We use the best performing model for the final submission in the evaluation phase. We provide details of the performance of the best performing model over the blind test dataset provided in the evaluation phase. We provide a detailed comparison of the performance of our models across all the class labels in the validation dataset in Table \ref{tab:results1}. Table \ref{tab:results1} shows that the simple BERT+Linear model (0.8758 F1 score) consistently performs better across all the labels (except for PROD) as compared to other larger models. We attribute this to the limited number of samples in the training dataset. The lack of a sufficient number of training samples limits the ability of larger models to generalize properly over the entire training set. 

Also, it can be observed from Table \ref{tab:results2} that all the 3 models outperform the baseline by a significant margin. BERT+CRF, BERT+BiLSTM+CRF, BERT+Linear advances the baseline by around 8\%, 6\%, and 9\% respectively. Table \ref{tab:results3} shows the performance of our BERT+Linear model on the blind test set. Our best performing model ranks 9th on the validation dataset and 15th on the final blind test set. Moreover, through our experiments, we find that the BERT-large offers a significant boost in performance over BERT-base, due to the larger number of attention heads.



\section{Error Analysis}
\label{erroranalysis}
We perform error analysis for all 3 different model performances on the validation dataset. We find that for all 3 models, each model has the greatest difficulty in accurately predicting the \textit{CW (Creative Work)} label. This can be attributed to the higher degree of ambiguity when it comes to \textit{CW} named entities, as these often share a similar type of textual structure as regular non-named entity text tokens. It can be inferred that all 3 models are memorizing entity names from the training data to some extent. It is most prevalent in BERT+BiLSTM+CRF model, as we can see that it has the least amount of prediction accuracy among other models. This is consistent with our reasoning that heavier models tend to overfit the dataset faster. Hence, we deduce that named entity memorization can be attributed to a type of overfitting behavior by the model in question when the training data is scarce. The BERT+Linear model, which is the lightest model with the least amount of trainable parameters among all 3, is found to be significantly less prone to memorize entity names.

Furthermore, upon qualitative analysis, we find that our models often have difficulty in recognizing longer named entities (entities comprising of 5 or more tokens). This can be attributed to the lack of occurrence of such entities in the training dataset.
The models are majorly exposed to shorter length entity spans across the training set. Due to the lack of exposure of the models to adequate training instances of longer spans, the models are often unable to predict such longer entity spans.

It is also worth noting that an increase in the number of attention heads in the BERT layer helps in substantial improvement in the accuracy. As discussed, this can be attributed to better learning of the context with the help of attention mechanism. We conclude that the larger number of attention heads are able to classify longer entity spans with greater accuracy.

\section{Conclusion and Future Work}
\label{conclusion}
We experiment with 3 model architectures for a novel dataset introduced for the shared task of detection of complex NER. Our best performing model comprises of a simple linear classifier on top of fine-tuned BERT-based language model. We find that this simple approach performs competitively as compared to its heavier counterparts. Upon analysis, we attribute this observation to the scarcity of labeled training data. BERT+Linear model is able to optimally avoid overfitting to a larger extent and hence performs better than other heavier models. We find that our simpler model ranks in the top 10 in the validation phase and outperforms numerous teams in the final evaluation phase. For future work, we aim to utilize other data augmentation techniques and distant supervision to create clean silver labels in order to increase our training instances. We believe that this would help us leverage larger models for training purposes.

\bibliography{anthology,custom}

\begin{thebibliography}{30}
\expandafter\ifx\csname natexlab\endcsname\relax\def\natexlab#1{#1}\fi

\bibitem[{Baevski et~al.(2019)Baevski, Edunov, Liu, Zettlemoyer, and
  Auli}]{baevski-etal-2019-cloze}
Alexei Baevski, Sergey Edunov, Yinhan Liu, Luke Zettlemoyer, and Michael Auli.
  2019.
\newblock \href {https://doi.org/10.18653/v1/D19-1539} {Cloze-driven
  pretraining of self-attention networks}.
\newblock In \emph{Proceedings of the 2019 Conference on Empirical Methods in
  Natural Language Processing and the 9th International Joint Conference on
  Natural Language Processing (EMNLP-IJCNLP)}, pages 5360--5369, Hong Kong,
  China. Association for Computational Linguistics.

\bibitem[{Devlin et~al.(2019)Devlin, Chang, Lee, and
  Toutanova}]{devlin2019bert}
Jacob Devlin, Ming-Wei Chang, Kenton Lee, and Kristina Toutanova. 2019.
\newblock \href {http://arxiv.org/abs/1810.04805} {Bert: Pre-training of deep
  bidirectional transformers for language understanding}.

\bibitem[{Florian et~al.(2003)Florian, Ittycheriah, Jing, and
  Zhang}]{florian2003named}
Radu Florian, Abe Ittycheriah, Hongyan Jing, and Tong Zhang. 2003.
\newblock Named entity recognition through classifier combination.
\newblock In \emph{Proceedings of the seventh conference on Natural language
  learning at HLT-NAACL 2003}, pages 168--171.

\bibitem[{Goldberg et~al.(2017)Goldberg, Wang, and Grant}]{10.1145/3012003}
Sean Goldberg, Daisy~Zhe Wang, and Christan Grant. 2017.
\newblock \href {https://doi.org/10.1145/3012003} {A probabilistically
  integrated system for crowd-assisted text labeling and extraction}.
\newblock \emph{J. Data and Information Quality}, 8(2).

\bibitem[{Habibi et~al.(2017)Habibi, Weber, Neves, Wiegandt, and
  Leser}]{habibi2017deep}
Maryam Habibi, Leon Weber, Mariana Neves, David~Luis Wiegandt, and Ulf Leser.
  2017.
\newblock Deep learning with word embeddings improves biomedical named entity
  recognition.
\newblock \emph{Bioinformatics}, 33(14):i37--i48.

\bibitem[{Hedderich et~al.(2021)Hedderich, Lange, and
  Klakow}]{DBLP:journals/corr/abs-2102-13129}
Michael~A. Hedderich, Lukas Lange, and Dietrich Klakow. 2021.
\newblock \href {http://arxiv.org/abs/2102.13129} {{ANEA:} distant supervision
  for low-resource named entity recognition}.
\newblock \emph{CoRR}, abs/2102.13129.

\bibitem[{Jain et~al.(2020)Jain, van Zuylen, Hajishirzi, and
  Beltagy}]{jain-etal-2020-scirex}
Sarthak Jain, Madeleine van Zuylen, Hannaneh Hajishirzi, and Iz~Beltagy. 2020.
\newblock \href {http://arxiv.org/abs/2005.00512} {Scirex: A challenge dataset
  for document-level information extraction}.
\newblock In \emph{Proceedings of the 58th Annual Meeting of the Association
  for Computational Linguistics}.

\bibitem[{Kocaman and Talby(2020)}]{biomed}
Veysel Kocaman and David Talby. 2020.
\newblock \href {http://arxiv.org/abs/2011.06315} {Biomedical named entity
  recognition at scale}.
\newblock \emph{CoRR}, abs/2011.06315.

\bibitem[{Lafferty et~al.(2001)Lafferty, McCallum, and
  Pereira}]{Lafferty2001ConditionalRF}
J.~Lafferty, A.~McCallum, and Fernando Pereira. 2001.
\newblock Conditional random fields: Probabilistic models for segmenting and
  labeling sequence data.
\newblock In \emph{ICML}.

\bibitem[{Lample et~al.(2016{\natexlab{a}})Lample, Ballesteros, Subramanian,
  Kawakami, and Dyer}]{lample2016neural}
Guillaume Lample, Miguel Ballesteros, Sandeep Subramanian, Kazuya Kawakami, and
  Chris Dyer. 2016{\natexlab{a}}.
\newblock Neural architectures for named entity recognition.
\newblock \emph{arXiv preprint arXiv:1603.01360}.

\bibitem[{Lample et~al.(2016{\natexlab{b}})Lample, Ballesteros, Subramanian,
  Kawakami, and Dyer}]{lample-etal-2016-neural}
Guillaume Lample, Miguel Ballesteros, Sandeep Subramanian, Kazuya Kawakami, and
  Chris Dyer. 2016{\natexlab{b}}.
\newblock \href {https://doi.org/10.18653/v1/N16-1030} {Neural architectures
  for named entity recognition}.
\newblock In \emph{Proceedings of the 2016 Conference of the North {A}merican
  Chapter of the Association for Computational Linguistics: Human Language
  Technologies}, pages 260--270, San Diego, California. Association for
  Computational Linguistics.

\bibitem[{Li et~al.(2018)Li, Sun, Han, and Li}]{Li2018ASO}
J.~Li, Aixin Sun, Jianglei Han, and Chenliang Li. 2018.
\newblock A survey on deep learning for named entity recognition.
\newblock \emph{ArXiv}, abs/1812.09449.

\bibitem[{Li et~al.(2020)Li, Sun, Han, and Li}]{li2020survey}
Jing Li, Aixin Sun, Jianglei Han, and Chenliang Li. 2020.
\newblock A survey on deep learning for named entity recognition.
\newblock \emph{IEEE Transactions on Knowledge and Data Engineering},
  34(1):50--70.

\bibitem[{Liang et~al.(2020)Liang, Yu, Jiang, Er, Wang, Zhao, and
  Zhang}]{DBLP:journals/corr/abs-2006-15509}
Chen Liang, Yue Yu, Haoming Jiang, Siawpeng Er, Ruijia Wang, Tuo Zhao, and Chao
  Zhang. 2020.
\newblock \href {http://arxiv.org/abs/2006.15509} {{BOND:} bert-assisted
  open-domain named entity recognition with distant supervision}.
\newblock \emph{CoRR}, abs/2006.15509.

\bibitem[{Luan et~al.(2018)Luan, He, Ostendorf, and
  Hajishirzi}]{luan-etal-2018-multi}
Yi~Luan, Luheng He, Mari Ostendorf, and Hannaneh Hajishirzi. 2018.
\newblock \href {https://doi.org/10.18653/v1/D18-1360} {Multi-task
  identification of entities, relations, and coreference for scientific
  knowledge graph construction}.
\newblock In \emph{Proceedings of the 2018 Conference on Empirical Methods in
  Natural Language Processing}, pages 3219--3232, Brussels, Belgium.
  Association for Computational Linguistics.

\bibitem[{Ma and Hovy(2016)}]{ma-hovy-2016-end}
Xuezhe Ma and Eduard Hovy. 2016.
\newblock \href {https://doi.org/10.18653/v1/P16-1101} {End-to-end sequence
  labeling via bi-directional {LSTM}-{CNN}s-{CRF}}.
\newblock In \emph{Proceedings of the 54th Annual Meeting of the Association
  for Computational Linguistics (Volume 1: Long Papers)}, pages 1064--1074,
  Berlin, Germany. Association for Computational Linguistics.

\bibitem[{Malmasi et~al.(2022{\natexlab{a}})Malmasi, Fang, Fetahu, Kar, and
  Rokhlenko}]{multiconer-data}
Shervin Malmasi, Anjie Fang, Besnik Fetahu, Sudipta Kar, and Oleg Rokhlenko.
  2022{\natexlab{a}}.
\newblock {MultiCoNER: a Large-scale Multilingual dataset for Complex Named
  Entity Recognition}.

\bibitem[{Malmasi et~al.(2022{\natexlab{b}})Malmasi, Fang, Fetahu, Kar, and
  Rokhlenko}]{multiconer-report}
Shervin Malmasi, Anjie Fang, Besnik Fetahu, Sudipta Kar, and Oleg Rokhlenko.
  2022{\natexlab{b}}.
\newblock {SemEval-2022 Task 11: Multilingual Complex Named Entity Recognition
  (MultiCoNER)}.
\newblock In \emph{Proceedings of the 16th International Workshop on Semantic
  Evaluation (SemEval-2022)}. Association for Computational Linguistics.

\bibitem[{Mansouri et~al.(2008)Mansouri, Affendey, and
  Mamat}]{mansouri2008named}
Alireza Mansouri, Lilly~Suriani Affendey, and Ali Mamat. 2008.
\newblock Named entity recognition approaches.
\newblock \emph{International Journal of Computer Science and Network
  Security}, 8(2):339--344.

\bibitem[{Mesbah et~al.(2018)Mesbah, Lofi, Torre, Bozzon, and
  Houben}]{mesbah2018tse}
Sepideh Mesbah, Christoph Lofi, Manuel~Valle Torre, Alessandro Bozzon, and
  Geert-Jan Houben. 2018.
\newblock Tse-ner: An iterative approach for long-tail entity extraction in
  scientific publications.
\newblock In \emph{International Semantic Web Conference}, pages 127--143.
  Springer.

\bibitem[{Mikolov et~al.(2013)Mikolov, Chen, Corrado, and
  Dean}]{mikolov2013efficient}
Tomas Mikolov, Kai Chen, Greg Corrado, and Jeffrey Dean. 2013.
\newblock Efficient estimation of word representations in vector space.
\newblock \emph{arXiv preprint arXiv:1301.3781}.

\bibitem[{Nadeau and Sekine(2007)}]{nadeau2007survey}
David Nadeau and Satoshi Sekine. 2007.
\newblock A survey of named entity recognition and classification.
\newblock \emph{Lingvisticae Investigationes}, 30(1):3--26.

\bibitem[{Nooralahzadeh et~al.(2019)Nooralahzadeh, L{\o}nning, and
  {\O}vrelid}]{nooralahzadeh-etal-2019-reinforcement}
Farhad Nooralahzadeh, Jan~Tore L{\o}nning, and Lilja {\O}vrelid. 2019.
\newblock \href {https://doi.org/10.18653/v1/D19-6125} {Reinforcement-based
  denoising of distantly supervised {NER} with partial annotation}.
\newblock In \emph{Proceedings of the 2nd Workshop on Deep Learning Approaches
  for Low-Resource NLP (DeepLo 2019)}, pages 225--233, Hong Kong, China.
  Association for Computational Linguistics.

\bibitem[{Pennington et~al.(2014)Pennington, Socher, and
  Manning}]{pennington-etal-2014-glove}
Jeffrey Pennington, Richard Socher, and Christopher Manning. 2014.
\newblock \href {https://doi.org/10.3115/v1/D14-1162} {{G}lo{V}e: Global
  vectors for word representation}.
\newblock In \emph{Proceedings of the 2014 Conference on Empirical Methods in
  Natural Language Processing ({EMNLP})}, pages 1532--1543, Doha, Qatar.
  Association for Computational Linguistics.

\bibitem[{Ramshaw and Marcus(1999)}]{ramshaw1999text}
Lance~A Ramshaw and Mitchell~P Marcus. 1999.
\newblock Text chunking using transformation-based learning.
\newblock In \emph{Natural language processing using very large corpora}, pages
  157--176. Springer.

\bibitem[{Ritter et~al.(2011)Ritter, Clark, Etzioni et~al.}]{ritter2011named}
Alan Ritter, Sam Clark, Oren Etzioni, et~al. 2011.
\newblock Named entity recognition in tweets: an experimental study.
\newblock In \emph{Proceedings of the 2011 conference on empirical methods in
  natural language processing}, pages 1524--1534.

\bibitem[{Vaswani et~al.(2017)Vaswani, Shazeer, Parmar, Uszkoreit, Jones,
  Gomez, Kaiser, and Polosukhin}]{vaswani2017attention}
Ashish Vaswani, Noam Shazeer, Niki Parmar, Jakob Uszkoreit, Llion Jones,
  Aidan~N. Gomez, Lukasz Kaiser, and Illia Polosukhin. 2017.
\newblock \href {http://arxiv.org/abs/1706.03762} {Attention is all you need}.

\bibitem[{Wang et~al.(2020)Wang, Guan, Zhang, Li, and Han}]{9378052}
Xuan Wang, Yingjun Guan, Yu~Zhang, Qi~Li, and Jiawei Han. 2020.
\newblock \href {https://doi.org/10.1109/BigData50022.2020.9378052}
  {Pattern-enhanced named entity recognition with distant supervision}.
\newblock In \emph{2020 IEEE International Conference on Big Data (Big Data)},
  pages 818--827.

\bibitem[{Yadav and Bethard(2019)}]{yadav2019survey}
Vikas Yadav and Steven Bethard. 2019.
\newblock A survey on recent advances in named entity recognition from deep
  learning models.
\newblock \emph{arXiv preprint arXiv:1910.11470}.

\bibitem[{Yang et~al.(2018)Yang, Chen, Li, He, and
  Zhang}]{yang-etal-2018-distantly}
Yaosheng Yang, Wenliang Chen, Zhenghua Li, Zhengqiu He, and Min Zhang. 2018.
\newblock \href {https://aclanthology.org/C18-1183} {Distantly supervised {NER}
  with partial annotation learning and reinforcement learning}.
\newblock In \emph{Proceedings of the 27th International Conference on
  Computational Linguistics}, pages 2159--2169, Santa Fe, New Mexico, USA.
  Association for Computational Linguistics.

\end{thebibliography}
\bibliographystyle{acl_natbib}

\end{document}